\documentclass[letterpaper]{article} 
\usepackage{aaai2026}  
\usepackage{times}  
\usepackage{helvet}  
\usepackage{courier}  
\usepackage[hyphens]{url}  
\usepackage{graphicx} 
\usepackage{natbib}  
\usepackage{caption} 
\usepackage{algorithm}
\usepackage{algorithmic}

\usepackage{booktabs}
\usepackage{amsmath}
\usepackage{multirow}
\usepackage{amssymb}
\usepackage{pifont}
\usepackage{subfiles}

\usepackage{newfloat}
\usepackage{listings}
\DeclareCaptionStyle{ruled}{labelfont=normalfont,labelsep=colon,strut=off} 
\floatstyle{ruled}
\newfloat{listing}{tb}{lst}{}
\floatname{listing}{Listing}
\nocopyright

\title{EmoVid: A Multimodal Emotion Video Dataset \\ for Emotion-Centric Video Understanding and Generation}

\author {
    Zongyang Qiu\textsuperscript{\rm 1,\rm 2}\equalcontrib,
    Bingyuan Wang\textsuperscript{\rm 1}\equalcontrib,
    Xingbei Chen\textsuperscript{\rm 1},
    Yingqing He\textsuperscript{\rm 3},
    Zeyu Wang\textsuperscript{\rm 1,\rm 3}\thanks{Corresponding author.}
}
\affiliations {
    \textsuperscript{\rm 1}The Hong Kong University of Science and Technology (Guangzhou), China\\
    \textsuperscript{\rm 2}Fudan University, China\\
    \textsuperscript{\rm 3}The Hong Kong University of Science and Technology, Hong Kong SAR, China\\
    zyqiu22@m.fudan.edu.cn, bwang667@connect.hkust-gz.edu.cn, xchen053@connect.hkust-gz.edu.cn, yhebm@connect.ust.hk, zeyuwang@ust.hk
}

\usepackage{bibentry}

\begin{document}

\maketitle

\begin{abstract}
Emotion plays a pivotal role in video-based expression, but existing video generation systems predominantly focus on low-level visual metrics while neglecting affective dimensions. Although emotion analysis has made progress in the visual domain, the video community lacks dedicated resources to bridge emotion understanding with generative tasks, particularly for stylized and non-realistic contexts.
To address this gap, we introduce \textbf{EmoVid}, the first multimodal, emotion-annotated video dataset specifically designed for artistic media, which includes cartoon animations, movie clips, and animated stickers. Each video is annotated with emotion labels, visual attributes (brightness, colorfulness, hue), and text captions. Through systematic analysis, we uncover spatial and temporal patterns linking visual features to emotional perceptions across diverse video forms. Building on these insights, we develop an emotion-conditioned video generation technique by fine-tuning the Wan2.1 model. The results show a significant improvement in both quantitative metrics and the visual quality of generated videos for text-to-video and image-to-video tasks.
EmoVid establishes a new benchmark and protocol for affective video computing. Our work not only offers valuable insights into visual emotion analysis in artistically styled videos, but also provides practical methods for enhancing emotional expression in video generation.
\end{abstract}

\begin{links}
    \link{Project Page}{https://zane-zyqiu.github.io/EmoVid}
\end{links}

\begin{figure*}[t]
    \centering
    \includegraphics[width=\textwidth]{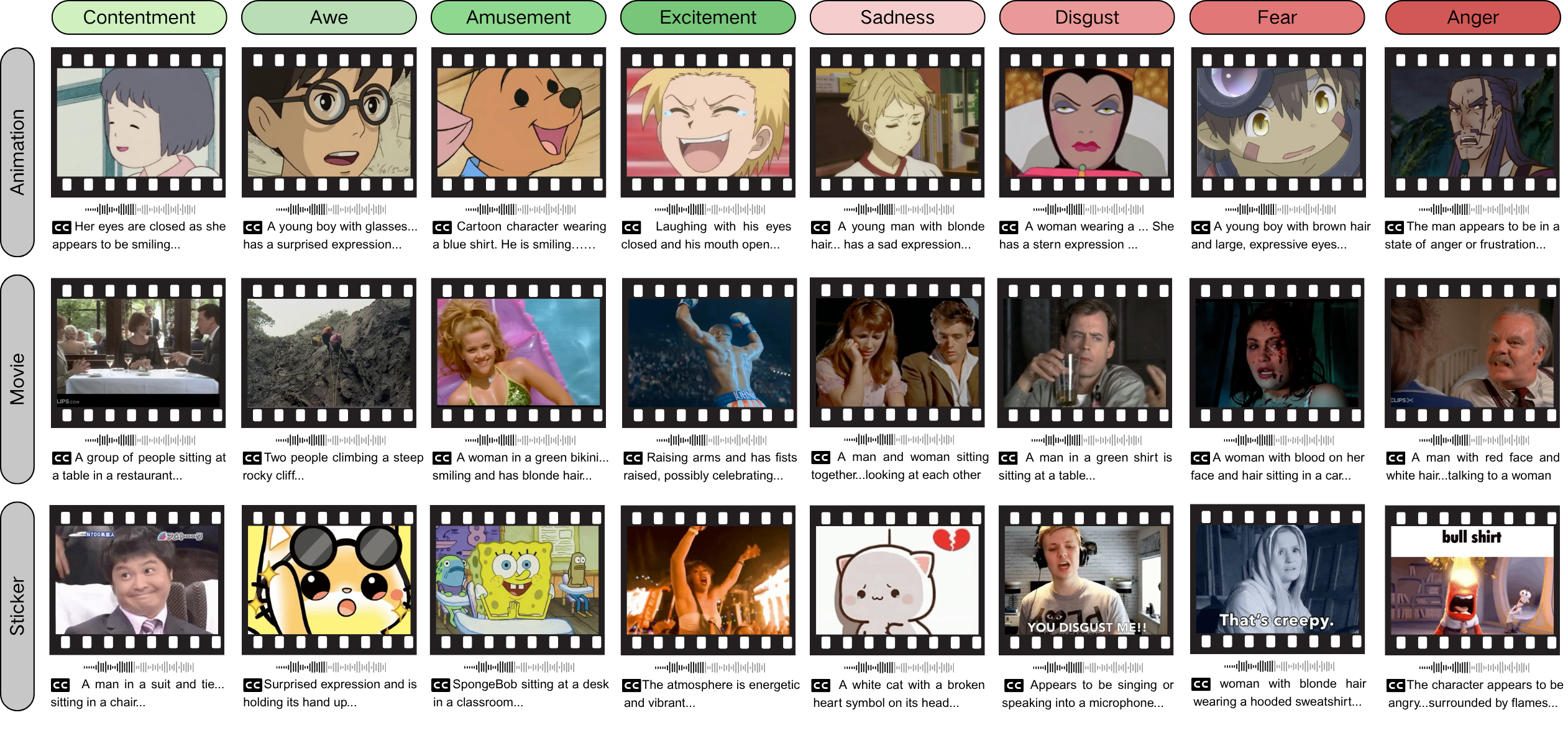}
    \caption{\textbf{Overview of the EmoVid dataset.} The dataset spans eight emotion categories—\textit{Contentment, Awe, Amusement, Excitement, Sadness, Disgust, Fear, and Anger}—and three content domains: \textit{Animation, Movie, and Sticker}. The dataset captures diverse emotional expressions in various visual styles and contexts, demonstrating both multimodal richness (with associated text and audio) and cross-domain generality.}
    \label{fig:overview}
\end{figure*}

\section{Introduction}

Video is a powerful medium for storytelling and expression, with emotion playing a key role in viewer engagement~\cite{cao2022video}. While recent video generation models have improved in visual coherence and motion, they have paid limited attention to emotional expressiveness~\cite{kalateh2024systematic}. This is especially true in creative applications like comic portrait animation, sticker (meme) creation, and cinematic editing, where emotional expressiveness is essential but underexplored~\cite{wang2025diffusion}.

In recent years, emotional content analysis and generation have achieved great progress in language, speech, and images, and have gained considerable attention within multimodal contexts~\cite{kalateh2024systematic}. In video, multimodal approaches have shown promising results in improving tasks such as sentiment analysis, emotion-driven content creation, and interactive video generation~\cite{pandeya2021deepmultimodal}. However, most of these studies focused on human dialogue and realistic styles. The integration of emotion in stylized video understanding and generation---taking into account both creative context, stylistic features, and the emotional undercurrents---remains underexplored.

In this paper, we introduce EmoVid, the first large-scale and emotion-labeled video dataset focusing on stylized and non-realistic content. As shown in Figure~\ref{fig:overview}, EmoVid consists of videos in three categories: cartoon animation, movie clips, and animated stickers (GIFs). We adopt the Mikels' eight-emotion scheme~\cite{mikels2005emotional} (amusement, awe, contentment, excitement, anger, disgust, fear, sadness), a widely used discrete set originally curated for affective image studies. Each clip is annotated with emotion labels, color attributes, and captioned via a vision-language model (VLM). Through this dataset, we explored the emotional patterns in videos, such as emotion distributions, color-emotion correlations, temporal transitions of emotion, and semantic links. We believe that these insights can enhance tasks like comic animation and stylized video generation.

To demonstrate the effectiveness of EmoVid, we propose a benchmark for both T2V and I2V tasks, evaluating the visual quality and emotion accuracy of AI-generated videos. We also fine-tune the Wan2.1~\cite{wan2025wan} model on our data. Experiments show a significant gain in emotional expressiveness when emotion is explicitly incorporated as a prior into video generation tasks. We also designed a video generation pipeline, which can generate animated stickers of any character with any emotion. This is of great value for today's network communication and can also be further used in the production of animations and movies. Together, EmoVid contributes to both affective computing and stylized video generation by linking emotion understanding with practical generative tasks.

In summary, we make the following contributions:
\begin{itemize}
    \item We introduce EmoVid, a large-scale, emotion-labeled video dataset focusing on stylized and non-realistic content, and present a scalable benchmark, evaluation metrics, and protocol for emotional enhancement in video generation.
    \item We explore both spatial and temporal emotional patterns in the EmoVid dataset, as well as their relationship with text captions or other visual attributes.
    \item We demonstrate EmoVid's utility in generation and editing tasks by fine-tuning the Wan2.1 model, which shows significant improvement in emotional expression.
\end{itemize}

\section{Related Work}

\begin{table*}[t]
\centering
\small
\setlength{\tabcolsep}{5pt}
\begin{tabular}{l l l l l}
\toprule
\textbf{Dataset} & \textbf{Modalities} & \textbf{Size} & \textbf{Content} & \textbf{Emotion Labels} \\
\midrule
CAER~\cite{lee2019context} & v, a & 12h (13k clips) & TV shows & 7 cls \\
MELD~\cite{poria2019meld} & v, a, t & 1.4h (1.4k clips) & Human dialogue & 7 cls \\
DEAP~\cite{koelstra2011deap} & v, a & 2h (120 clips) & Music videos & V-A-D \\
VEATIC~\cite{ren2024veatic} & v & 3h (124 clips) & In-the-wild & V-A \\
MEAD~\cite{wang2020mead} & v, a & 40h (10k clips) & Human face & 7 cls + 3 intensity levels \\
DH-FaceEmoVid-150~\cite{liu2025moee} & v, t & 150h (18k clips) & Human face & 6 cls + 4 compound \\
\textbf{EmoVid (Ours)} & \textbf{v, a, t} & \textbf{39h (22k clips)} & \textbf{Animation, Movie, Sticker} & \textbf{8 cls (Mikels)} \\
\bottomrule
\end{tabular}
\caption{\textbf{Comparison of EmoVid with other emotional video datasets.} We focus on modalities, size, content types, and emotion label schemes. 
Modalities are abbreviated as follows: \textit{v = Video, a = Audio, t = Text}. 
Emotion labels include either discrete categories (e.g., 7 cls = 7 emotion classes) or dimensional annotations such as \textit{Valence–Arousal (V–A)} and \textit{Valence–Arousal–Dominance (V–A–D)}. }
\label{tab:emotion_datasets}
\end{table*}

\subsection{Emotion Analysis and Affective Computing}
Affective computing has gained increasing attention in recent years, particularly in textual, auditory, and visual modalities. Early research mostly focused on text sentiment analysis using lexical features and semantic understanding~\cite{wang2025emotionlens}. Subsequently, auditory emotion recognition emerged as a reliable signal through speech prosody, pitch, and tone analysis~\cite{zadeh2016mosi}. Visual emotion recognition, especially via facial expression and gesture, has gained momentum more recently but remains less mature in comparison to text and audio~\cite{zhu2024review}. More recently, multimodal affective computing has become a key focus~\cite{das2023multimodal}, but comprehensive multimodal tasks remain scarce due to challenges in data alignment, modality imbalance, and dataset availability.

Within the visual modality, affective computing first progressed on static images, exploring the affective content of images via color histograms, facial attributes, and compositional cues~\cite{pang2015deep}. Nevertheless, static images lack temporal dynamics, which is essential for modeling transitions and nuances in affect. Consequently, video-based affective computing has gained traction, with datasets such as AffectNet and LIRIS-ACCEDE supporting sequence-based modeling~\cite{mollahosseini2017affectnet, baveye2015liris}. Despite progress, few efforts have been made to recognize video emotions within creative domains, such as film, performance, or storytelling, where affect is most deeply embedded and semantically rich. 

\subsection{Video Generation and Editing}
Recent advances in video generation have shown remarkable capability across diverse domains, such as human motion synthesis~\cite{tulyakov2018mocogan}, natural scene rendering~\cite{wang2018vid2vid}, and short video creation~\cite{ho2022imagen}. In creative applications---such as animation, film production, and meme/sticker creation---generative models like VideoCrafter~\cite{yang2023videocrafter} have demonstrated strong visual coherence and temporal smoothness, but current research primarily focuses on visual quality, realism, or aesthetic control, with little emphasis on affective expressiveness~\cite{ma2025controllable}. Notably, meme or sticker generation inherently carries affective signals, yet emotional intent is typically implicit and lacks formal integration into generation frameworks.

While affect has been discussed in video synthesis through domains such as facial expression transfer~\cite{zakharov2019few}, and gesture-guided animation~\cite{cao2022gesture}, these discussions are often limited to human-centric or conversational tasks. Affective conditioning has been explored via latent space alignment~\cite{ji2023emovideo} or emotion labels in prompt-based generation~\cite{guo2023animatediff}, yet these approaches rarely address stylized and non-realistic domains where emotion is central to narrative structure, such as animated films or cinematic scene generation~\cite{wang2025magicscroll}. The gap between affective modeling and creative video generation suggests a pressing need to bridge semantic emotion representation with generative visual storytelling.

\subsection{Emotion-related Datasets}

\label{sec:dataset}
Affective computing has been supported by a growing number of emotion datasets across textual, auditory, and visual modalities. Text-based corpora such as SemEval~\cite{rosenthal2017semeval} and GoEmotions~\cite{demszky2020goemotions} provide fine-grained emotion labels for sentiment and intent understanding. In the auditory domain, datasets like RAVDESS~\cite{livingstone2018ravdess} and IEMOCAP~\cite{busso2008iemocap} include speech with emotion expressions. Visual emotional datasets evolved from static-image datasets such as Emotion6~\cite{peng2015mixed} and EmoSet~\cite{yang2023emoset}. However, static images lack temporal continuity, limiting their utility in studying emotional dynamics and transitions. This has motivated the development of video-based datasets for affective modeling.


Several datasets have attempted to bridge the gap between affective labeling and video modality. MELD~\cite{poria2019meld}, DEAP~\cite{koelstra2011deap}, and VEATIC~\cite{ren2024veatic} offer emotion-annotated videos, but are limited in either modality (e.g., lacking audio and text), domain focus (e.g., only dialogues or music videos), or size. Facial datasets such as MEAD~\cite{wang2020mead} and DH-FaceEmoVid-150~\cite{liu2025moee} focus on constrained emotional expressions, offering high precision for facial affect but limited diversity in content and setting. These datasets are suitable for emotion recognition but not ideal for video generation, where varied visuals, rich context, and narrative emotional arcs are critical. In contrast, EmoVid introduces a large-scale, multimodal (video, audio, text) dataset with 22,758 clips covering animation, film scenes, and stickers---domains where emotional content is not only embedded but essential to semantics.

\section{The EmoVid Dataset}
As discussed in the former sections, the primary challenge of emotion-enhanced video generation lies in the lack of stylized emotional video datasets. We analyzed previous datasets specifically in emotion and video-related fields, and summarized their features in terms of size, category, content, and emotion labels, as shown in Table~\ref{tab:emotion_datasets}. Through the analyses, we find that existing datasets either lack in scale or fail to include all necessary modalities, which hinders the progress of multimodal emotion analysis. Moreover, they all focus exclusively on real-world scenarios (primarily human facial expressions), limiting the effective transfer of emotional priors to general video generation tasks.

\subsection{Data Collection}
To fill this gap, we created the EmoVid dataset, the first large-scale multimodal video dataset with fine-grained emotion labels, which comprises 2,807 animation face clips, 13,255 movie clips, and 6,696 animated stickers. The dataset includes high-quality emotional annotations, as well as other relevant visual attributes such as brightness, colorfulness, and hue, along with a textual caption for each video clip. Basic information is provided in Table~\ref{tab:emovid_stats}, and more detailed information can be found in the appendix.

\begin{table}[h]
\centering
\small
\begin{tabular}{lcccccc}
\toprule
\textbf{Type} & \textbf{Clips} & \textbf{Avg} & \textbf{SD} & \textbf{Vid} & \textbf{Aud} & \textbf{Cap} \\
\midrule
\textbf{Total}     & \textbf{22758} & \textbf{6.18} & \textbf{4.53} & -- & -- & -- \\
Animation &  2807 & 5.12 & 2.65 & \ding{51} & \ding{51} & \ding{51} \\
Movie     & 13255 & 8.75 & 4.71 & \ding{51} & \ding{51} & \ding{51} \\
Sticker   &  6696 & 2.91 & 2.18 & \ding{51} & \ding{55} & \ding{51} \\
\bottomrule
\end{tabular}
\caption{\textbf{Basic statistics of the EmoVid dataset.} \textit{Avg} is the average duration (in seconds), and \textit{SD} is the standard deviation. 
\textit{Vid}, \textit{Aud}, and \textit{Cap} indicate whether each clip includes video, audio, and textual caption, respectively. }
\label{tab:emovid_stats}
\end{table}

For the animation clips, we source data from the MagicAnime dataset, which contains 3,000 clips of cartoon faces from American, Chinese, and Japanese cartoons~\cite{xu2025magicanimehierarchicallyannotatedmultimodal}. The movie clips are retrieved using the metadata and code provided by Condensed Movies~\cite{bain2020condensed}. As movie videos are mostly several minutes long, we segment them using the PySceneDetect tool~\cite{Castellano_PySceneDetect}. Only clips within 4--30 seconds were retained for further analysis. For the animated stickers, we used the Tenor API~\cite{tenor2025} to search for GIFs based on the eight primary emotion labels and their synonyms summarized by~\citet{yang2023emoset}. Each clip is manually verified to ensure it accurately expresses the intended emotion. More details are included in the appendix.
\begin{figure}[b]
    \centering
    \includegraphics[width=\linewidth]{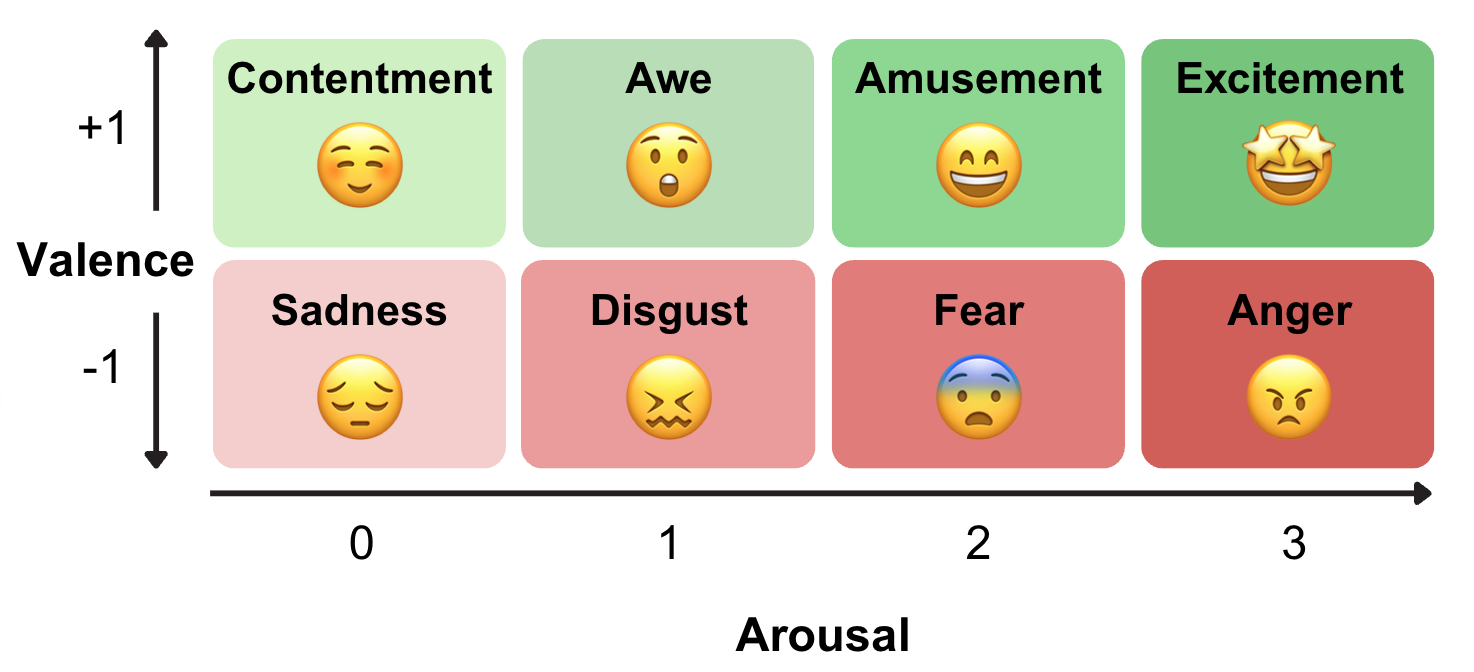}
    \caption{\textbf{Relationship between different emotions.} We refer to~\citet{warriner2013norms} to arrange emotion categories on the valence-arousal model.}
    \label{fig:V-A}
\end{figure}
\begin{figure}[b]
    \centering
    \includegraphics[width=\linewidth]{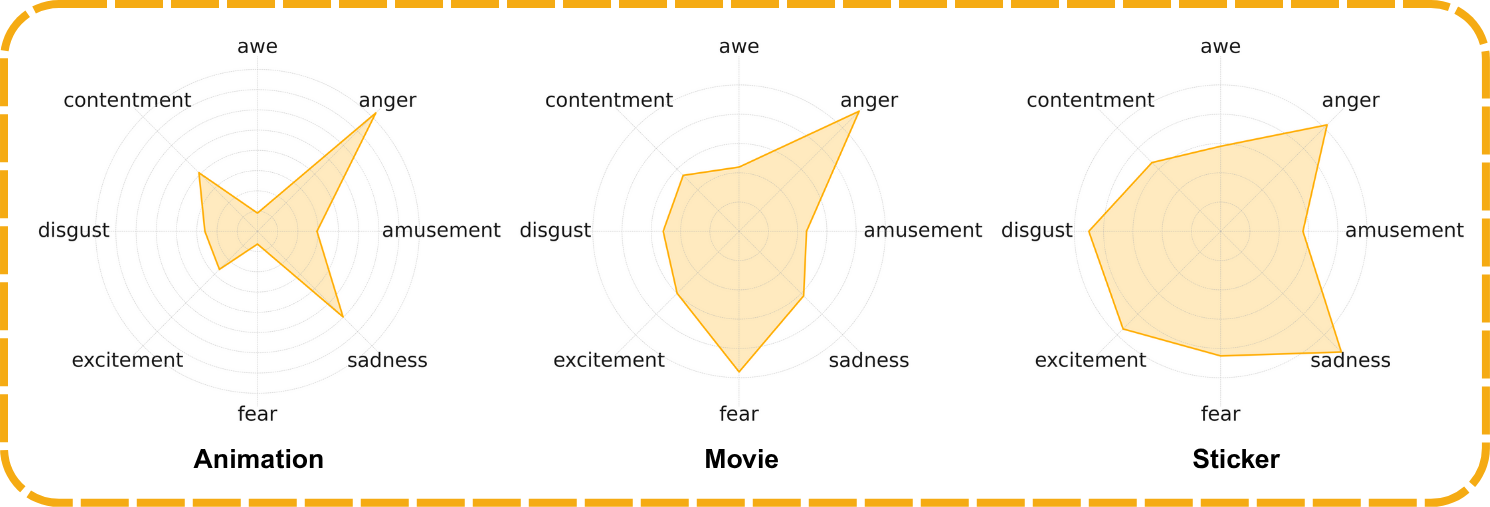}
    \caption{\textbf{Emotion distribution across three video categories.} Notably, the imbalance of \textit{animation} and \textit{movie} videos reflects the real-world emotional landscape of these domains.}
    \label{fig:radar}
\end{figure}
\subsection{Data Labeling}
\begin{figure*}[t]
    \centering
    \includegraphics[width=\linewidth]{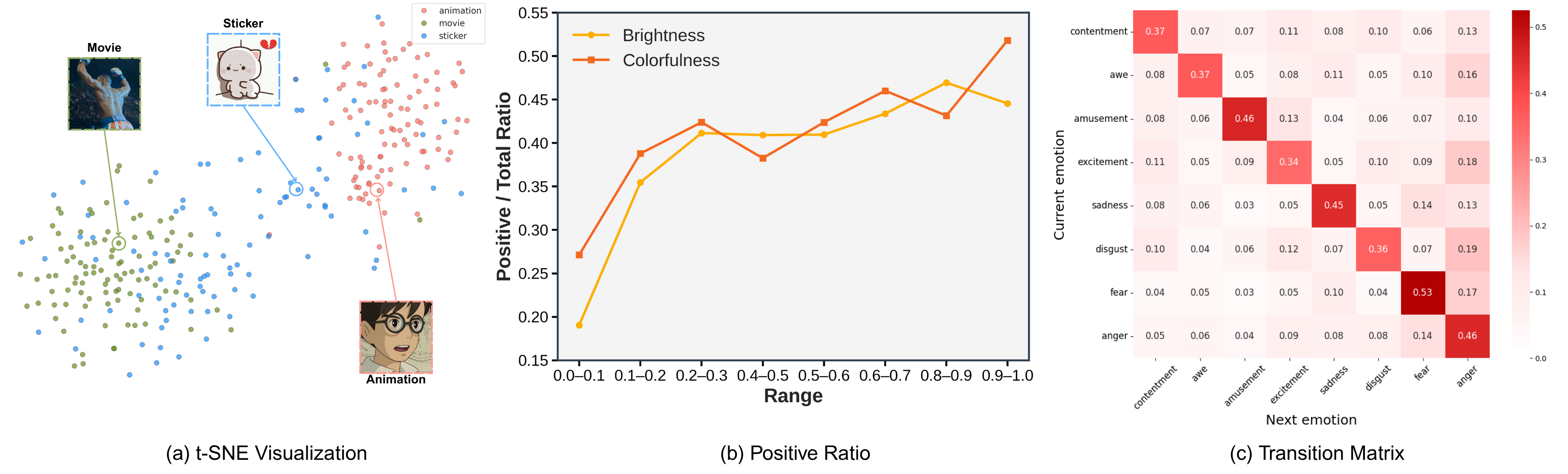}
    \caption{\textbf{Video features and color-emotion correlations.}
(a) t-SNE visualization of video features. Animation and Movie clusters are separated, with Sticker samples overlapping both, reflecting their hybrid content characteristics.
(b) Positive-to-total emotion ratio across bins of colorfulness and brightness, exhibiting a distinct upward trend.
(c) Emotion transition matrix from consecutive
movie clips. Diagonal dominance indicates strong emotional persistence.}
    \label{fig:color attributes}
\end{figure*}

Videos in EmoVid are annotated at the clip level, and the annotation includes the following aspects:

\textbf{Emotion.} We employ the widely-used Mikels emotion model~\cite{mikels2005emotional}, which categorizes emotions into eight types: \textit{amusement, awe, contentment, excitement, anger, disgust, fear, and sadness}. As the Valence-Arousal Model~\cite{russell1980circumplex} also plays an important role in intuitive understanding of emotions, we sorted the valence and arousal of the eight emotions as in Figure~\ref{fig:V-A} according to the work of~\citet{warriner2013norms}. 

Given the trade-off between labeling accuracy and resource consumption, we adopt a human-machine collaborative method to obtain the labels. We first conducted a comparative experiment on the EmoSet dataset~\cite{yang2023emoset}, the results of which are detailed in the appendix. We find that fine-tuning VLMs on the same data domain significantly improves emotion labeling accuracy, and NVILA-Lite-2B~\cite{liu2024nvila} exhibits classification performance comparable to that of humans. We pick 20\% of the animation and movie data to be annotated by human annotators, with each clip tagged as one of the eight emotions or as \textit{no specific emotion}. As emotions are ambiguous and open to interpretation, each video is annotated by three people, and the video is retained only when at least two annotators provide the same result. For the remaining 80\% of the data, we use the NVILA-Lite-2B model (fine-tuned on the manually labeled data) to annotate the clips. To assess annotation quality, we randomly select 1\% of the videos as a validation set. Three human annotators independently annotate the same set. We then calculated pairwise Cohen's kappa scores across the four annotation sources (three humans and the VLM). The results indicate a small difference ($< 4\%$) between the inter-human kappas and the human-VLM kappas, indicating that the VLM provides labels of similar quality to humans.

\textbf{Attributes and Captions.} We computed three low-level visual attributes for each video clip: \textit{colorfulness, brightness, and hue}, based on the HSV color space. Specifically, we sampled every 20 frames from each clip, and every pixel is represented by a triplet \((h_i, s_i, v_i)\). For hue \(H\), each pixel value \(h_i\) 
is defined as an angle on the color wheel in the range \([0,360)\), and the overall \(H\) is defined as the angle of the vector resulting from the sum of these vectors:

\begin{align}
\phi &= \operatorname{atan2}\left( \sum_{i} \sin(h_i), \sum_{i} \cos(h_i) \right) \times \frac{180^\circ}{\pi} \\
H &= \text{round}\left((\phi + 360^\circ) \bmod 360^\circ\right)
\end{align}

The modulo operation ensures that \(H\) lies within the range \([0,360)\). Colorfulness (\(C\)) was defined as the normalized average of the saturation channel (\(S\)), and brightness (\(B\)) was calculated using the value channel (\(V\)):
\begin{equation}
C = \text{round}\left(\frac{1}{N}\sum_{i=1}^{N} S_i, 1\right),B = \text{round}\left(\frac{1}{N}\sum_{i=1}^{N} V_i, 1\right).
\end{equation}
where \(N\) is the total number of pixels sampled from selected frames. Both \(C\) and \(B\) were normalized between 0 and 1 and rounded to one decimal place.

Additionally, we generated high-quality captions for each clip using the NVILA-8B-Video model to facilitate further training and evaluation based on the dataset.

\vspace{0.5em}
To sum up, beyond categorical emotion labels, EmoVid provides rich multimodal annotations including:
\begin{itemize}
  \item \textbf{Audio tracks} aligned with each video to enable audio-visual emotion fusion.
  \item \textbf{Low-level visual features} including \textit{brightness}, \textit{colorfulness}, and \textit{hue}, quantitatively extracted to support emotion attribution analysis.
  \item \textbf{Free-form captions} generated by a VLM, describing the perceived content and sentiment of each video.

\end{itemize}

\section{Analysis of EmoVid}

\subsection{Properties of EmoVid}

EmoVid is a large-scale, multimodal video dataset designed for emotion-aware video understanding and generation tasks. The dataset consists of 22,758 videos, with a total duration of 140,580 seconds. Among them, 10,049 clips have human-annotated emotion labels (282 animation clips, 2,771 movie clips, and 6,996 sticker clips). The emotion labels of the rest are generated by a fine-tuned VLM, the quality of which is verified in the above experiments.

As depicted in Figure~\ref{fig:color attributes}(a), the t-SNE visualization illustrates the clear distribution among the three data types. We observe significant differentiations between Animation and Movie types, while Sticker data points are intermediate, exhibiting overlaps with both Animation and Movie clusters. This aligns intuitively with expectations regarding content similarity across these categories.

EmoVid covers eight discrete emotion categories aligned with the Mikels model, including \textit{amusement, anger, awe, contentment, disgust, excitement, fear}, and \textit{sadness}. The videos are sourced from three representative domains---\textit{animation}, \textit{movie}, and \textit{sticker} content---each contributing different emotional intensities, stylistic traits, and temporal structures. Figure~\ref{fig:radar} summarizes the distribution of emotion labels across domains. From the figure, we observe that the emotion distribution in animation and movie video types is relatively imbalanced, with emotions like anger and sadness appearing more frequently, while amusement and awe are underrepresented. This reflects the natural distribution of emotions in real-life settings because these videos are all collected from real-world contexts and annotated.

Together, these properties make EmoVid an interpretable and extensible benchmark for video-based emotion understanding and generation. The dataset serves as a foundation for multimodal tasks such as emotional storytelling, text-to-video (T2V) or image-to-video (I2V) generation, and emotionally grounded video editing.

\begin{table*}[t]
\centering
\small
\setlength{\tabcolsep}{10pt}
\begin{tabular}{llcccccc}
\toprule
\textbf{Task} & \textbf{Method} & \textbf{FVD}↓ & \textbf{CLIP}↑ & \textbf{SD}↑ & \textbf{Flicker}↓ & \textbf{EA-2cls}↑ & \textbf{EA-8cls}↑ \\
\midrule
\multirow{5}{*}{T2V} 
  & VideoCrafter-V2           & 610.1 & 0.3012 & --   & 0.0184	 & 80.42 & 42.50\\
  & HunyuanVideo              & \textbf{552.6} & 0.2776 & -- & \textbf{0.0116} & 76.87 & 40.41 \\
  & CogVideoX                & 584.0 & 0.3013 & -- & 0.0213 & 82.91 & 44.58 \\
  & WanVideo (before)    & 594.3 & 0.2982 & -- & 0.0091 & 84.17 & 44.16 \\
  & WanVideo (after)     & 573.7 & \textbf{0.3021} & -- & 0.0143 & \textbf{88.33} & \textbf{48.33} \\
\midrule
\multirow{5}{*}{I2V} 
  & DynamiCrafter512          & \textbf{512.3} & -- & \textbf{0.7288} & 0.0280 & 90.41 & 71.25 \\
  & HunyuanVideo              & 544.6 & -- & 0.7244 & \textbf{0.0233} & 89.17 & 70.00 \\
  & CogVideoX                 &  528.4 & -- & 0.7214  & 0.0331 & 90.83 & 70.83 \\
  & WanVideo (before)     & 517.9 & -- & 0.7146 & 0.0325 & 91.25 & 71.30 \\
  & WanVideo (after)      & 517.8 & -- & 0.7193 & 0.0324 & \textbf{94.58} & \textbf{76.25} \\
\bottomrule
\end{tabular}
\caption{\textbf{Quantitative results on the EmoVid benchmark.} Evaluation covers T2V and I2V tasks. Emotion Accuracy (EA) is assessed using binary (EA-2cls) and 8-category (EA-8cls) emotion classification metrics, highlighting the improved capability of finetuned models in capturing and reproducing emotional content.}
\label{tab:experiment_results}
\end{table*}

\subsection{Emotional Structure and Dynamics}

\textbf{Visual attributes.} The inclusion of visual attributes in the dataset is intended to facilitate further research on the relationship between emotion and color expression. In Figure~\ref{fig:color attributes}(b), the positive-to-total emotion ratio demonstrates a general upward trend concerning colorfulness and brightness attributes, which is consistent with our conventional knowledge. We also calculate the average value of colorfulness, brightness, and hue for each of the eight emotions. Positive-valence categories are brighter and slightly more colorful than negative-valence ones, while high-arousal emotions tend to be darker but more colorful than low-arousal emotions. ANOVA~\cite{st1989analysis} confirms statistically significant ($p<0.01$) but small effects ($\eta^2 < 1\%$), and the details are in the appendix.


\textbf{Temporal analysis}. Different from static data, a key advantage of video datasets lies in the additional temporal dimension, which enables the exploration of how emotions evolve over time. Since the movie clips are extracted from continuous segments of films, we are able to perform temporal analysis based on them. As illustrated in Figure~\ref{fig:color attributes}(c), the first-order Markov transition matrix derived from consecutive movie clips unveils a three-stage emotional dynamic. First, all eight emotions exhibit strong self-persistence---particularly fear (0.53), anger (0.46), and amusement (0.46)---indicating that once an affective state is established, the visual stream tends to maintain it over short temporal windows. Second, transitions are markedly more frequent within the same valence polarity than across it (typically $0.08\text{--}0.18$ versus $<0.08$). Third, negative emotions reveal a chain-like escalation pattern: sadness \textrightarrow fear/anger and fear \textrightarrow anger---suggesting a possible ``defense-attack'' progression that makes negative sequences harder to dissolve~\cite{blanchard1977attack}. Together, these findings corroborate a ``hold, intra-valence drift, arousal leap'' trajectory, providing explicit understandings for emotion-aware video generation, editing, and pacing strategies.

\textbf{Text Caption.} To further explore the relationship between content and emotion, we extracted the five most frequently occurring 2–4 word phrases from video captions associated with each emotion, after filtering out redundant or noisy expressions. These phrases provide concrete semantic cues reflective of emotional content. For instance, under \textit{amusement} we found frequent phrases such as ``funny reaction,'' ``merry moment,'' and ``laughing together.'' In contrast, \textit{fear} often included phrases like ``dark tunnel,'' ``screaming sound,'' and ``approaching danger.'' The detailed information can be found in the appendix.

\section{Evaluation of EmoVid}

To rigorously evaluate the effectiveness and utility of EmoVid, we construct a comprehensive benchmark and perform both quantitative and qualitative analyses. Specifically, we sample 240 videos across 3 video types with 8 distinct emotion labels, selecting 10 representative videos for each category. The test videos are all sampled from human-annotated ones and have been double-checked to guarantee the best quality. For each video, we use its corresponding caption and modify it by appending an explicit emotional label (e.g., \textit{The video is in the ``amusement'' emotion}). 

We test four SOTA T2V models---VideoCrafter-V2~\cite{chen2024videocrafter2}, HunyuanVideo~\cite{kong2024hunyuanvideo}, CogVideoX-5B~\cite{yang2024cogvideox}, and Wan2.1-T2V-14B~\cite{wan2025wan}.
We also conduct experiments on four I2V models---DynamiCrafter512~\cite{xing2023dynamicrafter}, HunyuanVideo-I2V, CogVideoX-I2V, and Wan2.1-I2V-480P. 

Furthermore, we fine-tune both T2V and I2V Wan2.1 models on EmoVid, excluding the data in the benchmark with the LoRA technique~\cite{hu2022lora}. We conducted fine-tuning using the DiffSynth Studio framework~\cite{diffsynth-studio} on an H20 GPU with 96 GB memory. To balance the distribution of training data, we did not use the entire set of movie clips. Instead, the final training dataset consisted of 2,727 animation clips, 8,000 movie clips, and 6,616 sticker clips. The LoRA configuration was set with rank=32, learning\_ rate=1e-4, training\_epoch=3, and batch\_size=1, the same as default settings in DiffSynth Studio.

\subsection{Quantitative Results}
\begin{figure*}[t]
    \centering
    \includegraphics[width=0.95\linewidth]{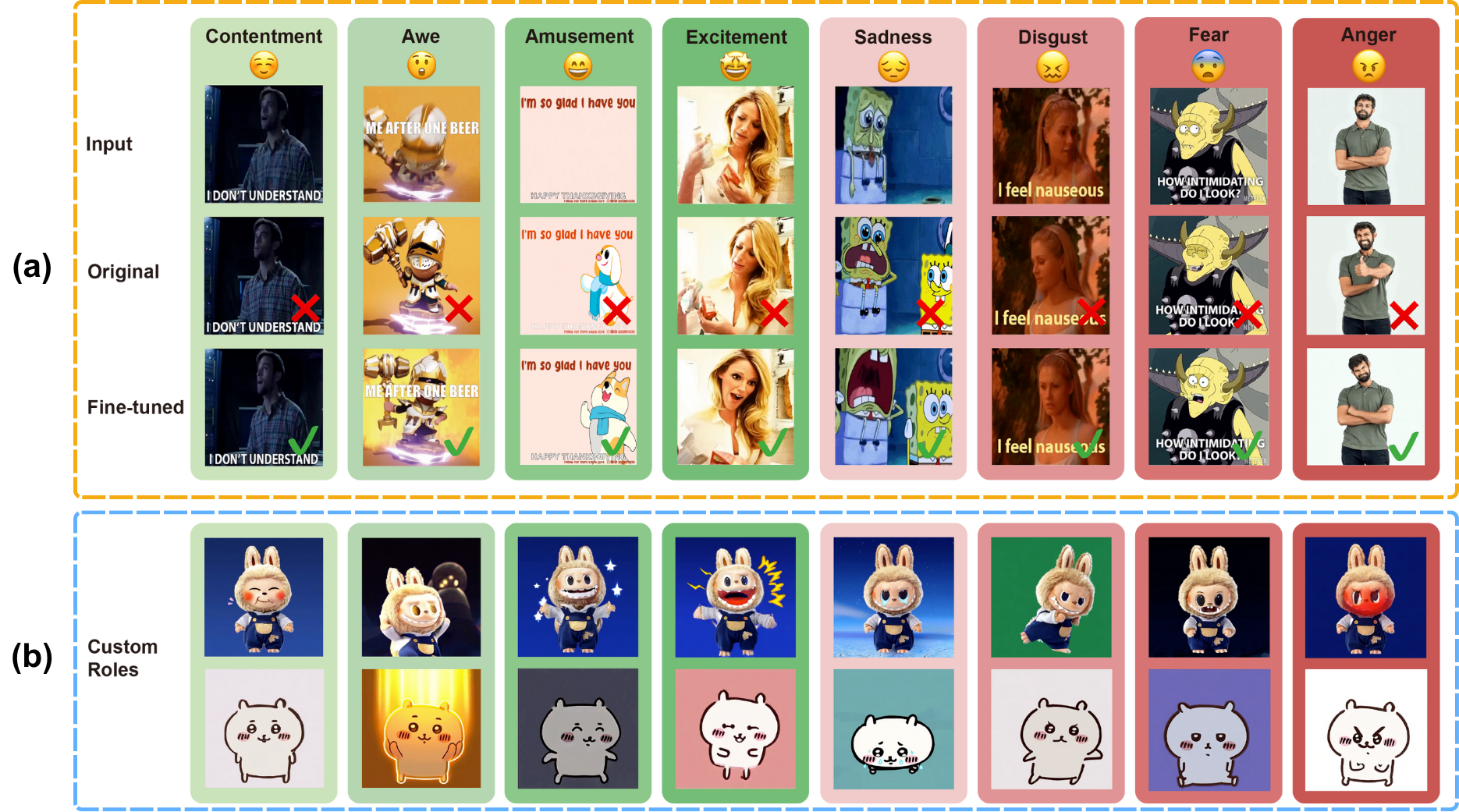}
    \caption{\textbf{Qualitative results.} 
(a) Comparison between the original Wan2.1 I2V model and our fine-tuned one. The \checkmark indicates better emotional alignment. 
(b) Emotion-conditioned animated sticker generation using the fine-tuned Wan2.1 I2V model.
 }
    \label{fig:qualitative results}
\end{figure*}

To measure the emotional accuracy of generated videos, we use the fine-tuned VLM employed during dataset annotation to classify the generated videos and adopt the same metrics as~\citet{yang2023emoset}. \textbf{EA-2cls} measures binary accuracy by checking whether the predicted emotion matches the ground-truth valence, while \textbf{EA-8cls} measures top-1 accuracy across the eight discrete emotions. We evaluated model performance using a diverse set of metrics: 
\begin{itemize}
    \item \textbf{FVD}~\cite{ge2024content} measures the overall visual fidelity.
    \item \textbf{CLIP Score}~\cite{hessel2021clipscore} quantifies semantic alignment between text and video.
    \item \textbf{SD Score}~\cite{liu2024evalcrafter} measures consistency between video and its first frame.
    \item \textbf{Temporal Flicker}~\cite{huang2024vbench} captures temporal instability across frames.
    \item \textbf{EA-2cls} and \textbf{EA-8cls} measure binary and full-class emotion accuracy.
\end{itemize}

Table~\ref{tab:experiment_results} shows that our fine-tuned model \textit{WanVideo (after)} has better performance in both tasks over the baseline \textit{WanVideo (before)}. In the T2V setting, while general metrics like FVD and CLIP show comparative performance, the emotion alignment metrics (EA-2cls and EA-8cls) exhibit clear gains, indicating stronger emotional expression fidelity in the generated videos. Similar trends are observed in the I2V scenario, where the fine-tuned model surpasses all competitors, achieving the highest emotion classification accuracy of \textbf{92.08\%} (2-class) and \textbf{72.92\%} (8-class).

\subsection{Qualitative Results}

To further assess the effectiveness of fine-tuning pretrained models on our dataset, we qualitatively compare video outputs of the original Wan2.1-I2V-480P model and those of the fine-tuned version. As shown in Figure~\ref{fig:qualitative results}(a), the baseline model often fails to capture the correct emotional tone---e.g., generating neutral or mismatched expressions---whereas the fine-tuned model exhibits more precise emotional articulation, such as heightened facial expressions, contextual cues, and mood-consistent motion patterns. These improvements highlight the value of EmoVid not only as a benchmarking dataset, but also as a valuable resource for emotion-specific downstream tasks.

We also utilize the LoRA-trained I2V model to generate animated stickers conditioned on different characters and emotions, as illustrated in Figure~\ref{fig:qualitative results}(b). The results demonstrate that our model is capable of producing vivid emotional expressions, which can be applied to social media platforms. In addition, we employ a multi-LoRA approach on the T2V model, combining our emotion-aware LoRA with other LoRAs that encode character identity or visual style priors, to generate videos with specific emotional attributes. More results can be found in the appendix.

\subsection{Discussion}
Through comparative experiments, we validate the effectiveness of the EmoVid benchmark. The quantitative results demonstrate that models fine-tuned with EmoVid data exhibit superior emotional accuracy, and the qualitative comparisons further support this conclusion. Such improvement mainly comes from cases where the baseline model failed to express the intended emotion, while the fine-tuned version captured it more precisely. Importantly, EmoVid is designed for artistic and creative scenarios where emotional expression is central. Such contexts include films and operas, miniseries, and emotionally evocative social media content such as expressive memes. In these applications, emotional clarity is often more important than realism or temporal fidelity. Our benchmark captures this focus by emphasizing emotional accuracy as the primary evaluation criterion.

Finally, we anticipate that EmoVid will be useful far beyond the scope of model benchmarking. Potential downstream applications include emotion-aware avatar generation, expressive media content synthesis, and controllable video editing based on emotional cues. As generative models become increasingly capable, datasets like EmoVid can help ground their outputs in meaningful human affect.

\section{Conclusion}
This paper has introduced \textbf{EmoVid}, the first large-scale emotion-annotated video dataset tailored specifically for creative contexts, including animations, movie clips, and animated stickers. EmoVid fills a critical gap by providing high-quality multimodal annotations—emotion labels, visual attributes, and textual captions—enabling deeper analysis of the interplay between visual features, temporal dynamics, and emotional perception. Through extensive experiments, we demonstrated EmoVid's capability in fine-tuning state-of-the-art generative models, resulting in significant improvements in emotional expressiveness for both T2V and I2V tasks.

By establishing a new benchmark for affective video computing, EmoVid not only advances fundamental research in emotion-driven video understanding and generation but also supports practical applications in fields such as animation, filmmaking, and social media communication. Our work is based on the assumption that each clip conveys a specific emotion. However, due to the complex nature of human emotion, the real-world expressions can be highly detailed and composite. In addition, the audio component of the dataset can be better leveraged to build a truly unified video-audio-text multimodal model. We will continue to explore these directions in our future work. 

\section{Ethics Statement} In recognition of the importance of copyright and privacy protection, we will only provide access to our dataset strictly for non-commercial research use by academic institutions. Any use of the dataset must comply with relevant intellectual property laws and ethical research standards. Redistribution or commercial use is prohibited.
\section{Acknowledgments}
We thank Shuolin Xu and Dr. Xian Xu for helpful discussions and support on dataset preprocessing. We also thank Xiaochun Wang for preparing the figures and Yihan Wu for assistance in conducting the user study.

\bibliography{aaai2026}

\clearpage

\begin{center}
    \Huge \bfseries APPENDIX
\end{center}

\section{Detailed Statistics}
The following section provides additional statistical insights and detailed analyses of the EmoVid dataset, further complementing the main manuscript.

\subsection{Duration Distribution across Styles}
\begin{figure}[b]
    \centering
    \includegraphics[width=1\linewidth]{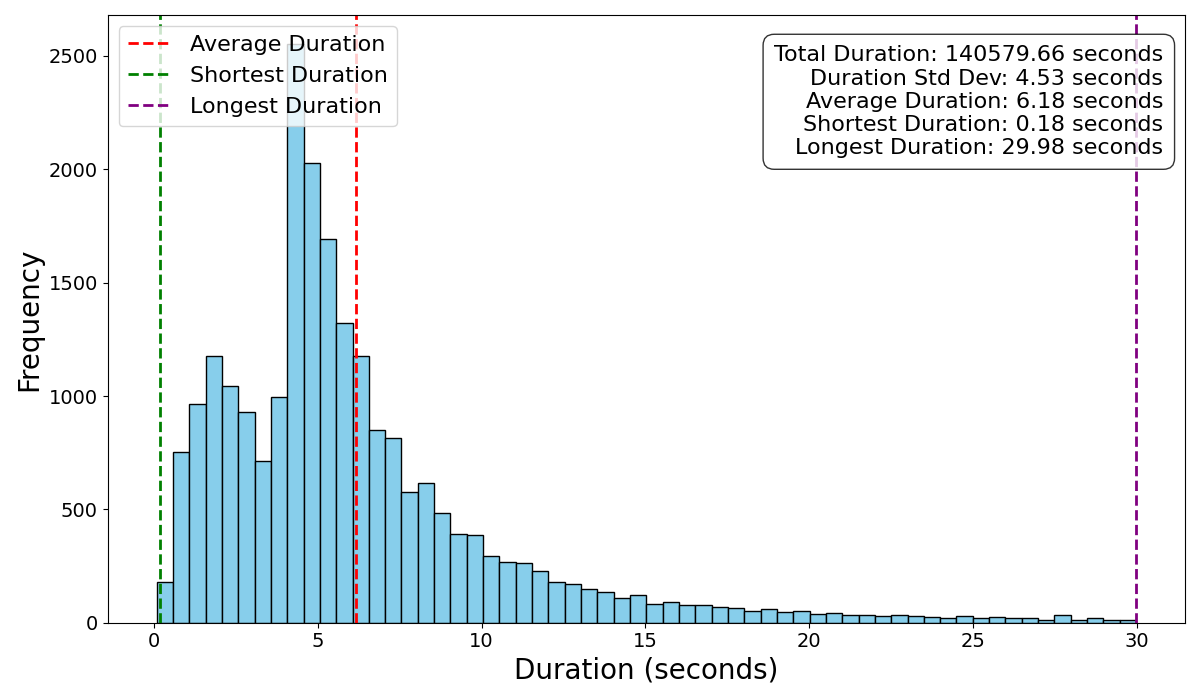}
    \caption{\textbf{Duration distribution of EmoVid clips.} Most clips cluster between roughly 2--8 seconds, and overall total duration is 140{,}580 seconds. Mean clip length is 6.18 seconds (min 0.18 s, max 29.98 s).}
    \label{fig:duration_distribution}
\end{figure}

Figure~\ref{fig:duration_distribution} illustrates the duration distribution of videos in the EmoVid dataset. The dataset encompasses a total duration of 140,580 seconds with a mean duration of 6.18 seconds per clip. The shortest video clip is 0.18 seconds, and the longest is 29.98 seconds. Most videos are concentrated between approximately 2 to 8 seconds, supporting both the capture of brief emotional moments and the exploration of extended emotional arcs.

\begin{figure}[b]
    \centering
    \includegraphics[width=1\linewidth]{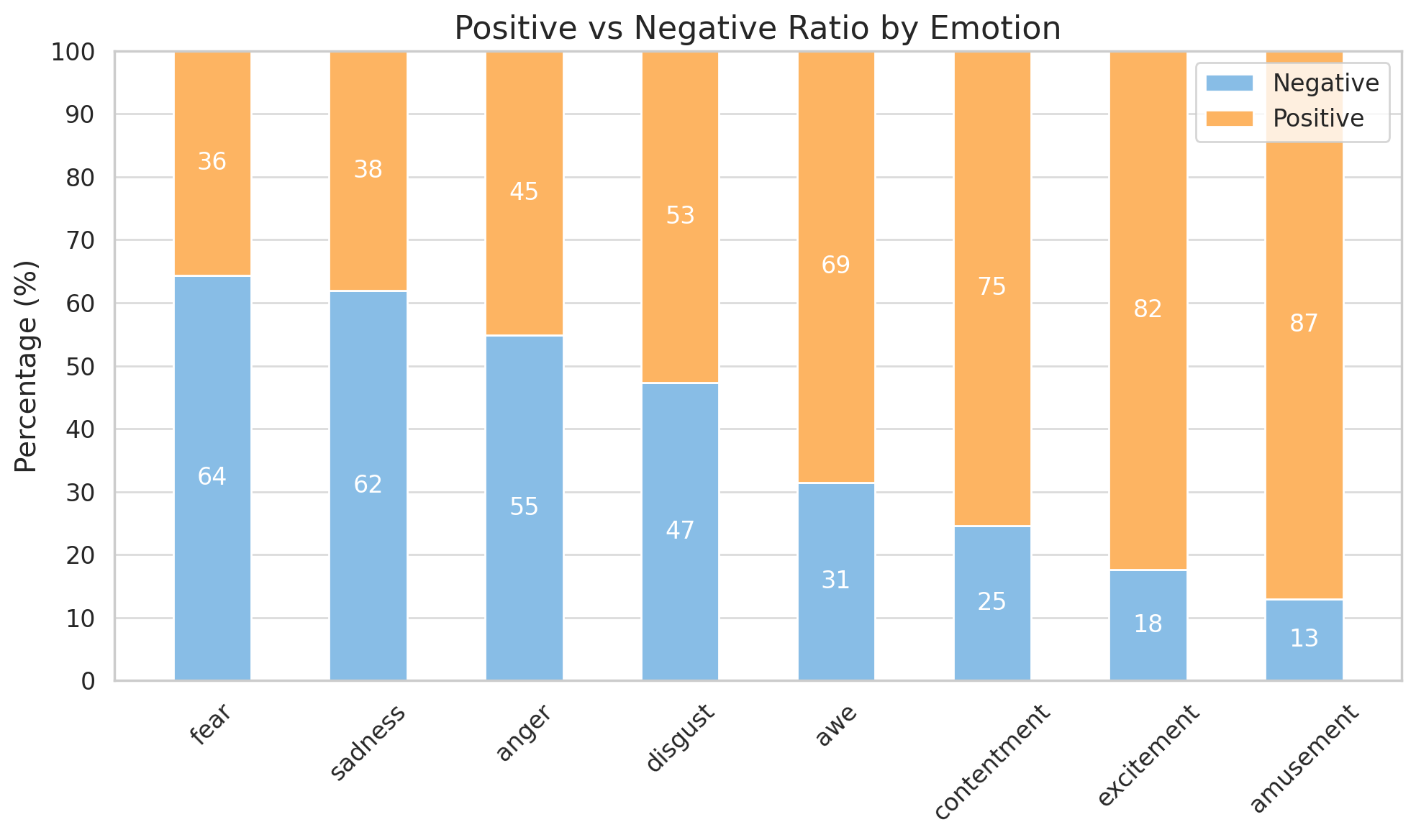}
    \caption{\textbf{Caption polarity by emotion category}. The results indicate good alignment between caption semantics and emotion labels.}
    \label{fig:pos/neg}
\end{figure}

Table~\ref{tab:emotion_distribution} provides a breakdown of emotion categories across different video styles---Animation, Movie, and Sticker. Movies constitute the largest share of data (58.24\%) and have the longest average video duration (8.75 seconds), followed by Stickers (29.42\%) with shorter clips (average 2.91 seconds) and Animations (12.33\%) averaging 5.12 seconds. This distribution enables comprehensive coverage of emotional expression across varying content styles.

\begin{table*}[t]
\centering
\setlength{\tabcolsep}{4pt}
\begin{tabular}{lrrrrrrrrrrr}
\toprule
\textbf{Style} & \textbf{Amusement} & \textbf{Anger} & \textbf{Awe} & \textbf{Contentment} & \textbf{Disgust} & \textbf{Excitement} & \textbf{Fear} & \textbf{Sadness} & \textbf{Total} & \textbf{Avg Len} & \textbf{SD}\\
\midrule
\textbf{Total} & \textbf{2008} & \textbf{4756} & \textbf{1768} & \textbf{2424} & \textbf{2458} & \textbf{2706} & \textbf{3315} & \textbf{3323} & \textbf{22758} & \textbf{6.18} & \textbf{4.53}\\
\midrule
Animation & 294 & 827 & 90 & 409 & 260 & 266 & 63 & 598 & 2807 & 5.12 & 2.65\\ 
Movie     & 1152 & 2901 & 1097 & 1351 & 1297 & 1497 & 2401 & 1559 & 13255 & 8.75 & 4.71\\
Sticker   & 562 & 1028 & 581 & 664 & 901 & 943 & 851 & 1166 & 6696 & 2.91 & 2.18\\
\bottomrule
\end{tabular}
\caption{\textbf{Distribution of emotions across styles.} \emph{Avg Len} denotes the mean clip duration in seconds; \emph{SD} is the standard deviation of clip length.}
\label{tab:emotion_distribution}
\end{table*}

\subsection{Color Attributes}

\begin{figure}[t]
    \centering
    \includegraphics[width=1\linewidth]{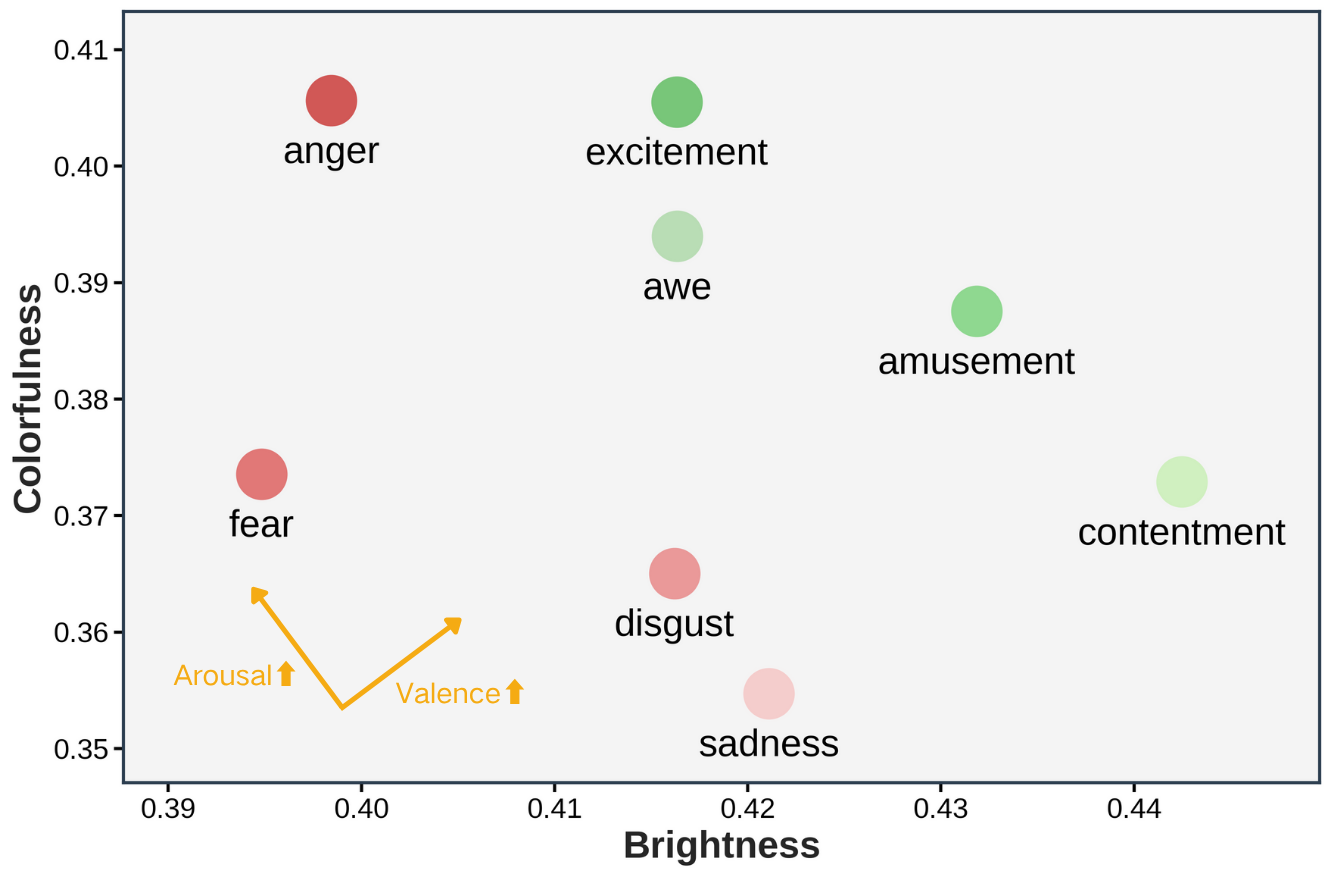}
    \caption{\textbf{Distribution of emotion categories in color space.}
Each point represents an emotion category, plotted by average brightness (x-axis) and colorfulness (y-axis).}
    \label{fig:color center}
\end{figure}

To explore the relationship between color attributes and video emotions, we calculated the mean and standard deviation of each color parameter for every emotion category, displayed in Table~\ref{tab:emotion_color_stats}. To properly handle the circular nature of hue values (ranging from $0^\circ$ to $360^\circ$), we compute the circular mean and circular variance. We observe that positive-valence emotions are generally brighter and more colorful, while high-arousal emotions tend to be darker but also more colorful than their low-arousal counterparts. ANOVA reveals statistically significant yet small effects ($\eta^2 < 1\%$), suggesting color cues are helpful for stylistic guidance or weak supervision, though insufficient for standalone classification. Figure~\ref{fig:color center} presents average values of colorfulness and brightness for each of the eight emotions. Notably, the spatial layout aligns well with valence and arousal dimensions, which form near-orthogonal directions, supporting the VA model’s robustness and highlighting a clear link between perceived emotion and color properties.

\begin{table}[t]
\centering
\small
\begin{tabular}{lccc}
\toprule
\textbf{Emotion} & \textbf{Brightness} & \textbf{Colorfulness} & \textbf{Hue (°)} \\
\midrule
Contentment & 0.443 ± 0.212 & 0.373 ± 0.195 & 34.50 ± 61.34 \\
Awe         & 0.416 ± 0.193 & 0.394 ± 0.199 & 38.47 ± 57.37 \\
Amusement   & 0.432 ± 0.208 & 0.388 ± 0.197 & 32.92 ± 54.42 \\
Excitement  & 0.416 ± 0.202 & 0.406 ± 0.200 & 38.21 ± 60.12 \\
Sadness     & 0.421 ± 0.211 & 0.355 ± 0.200 & 33.98 ± 64.69 \\
Fear        & 0.395 ± 0.205 & 0.374 ± 0.197 & 38.31 ± 60.23 \\
Disgust     & 0.416 ± 0.198 & 0.365 ± 0.192 & 43.06 ± 59.60 \\
Anger       & 0.398 ± 0.186 & 0.406 ± 0.198 & 37.74 ± 52.05 \\
\bottomrule
\end{tabular}
\caption{\textbf{Per-emotion visual feature analysis.} We include statistics of brightness, colorfulness, and hue.}
\label{tab:emotion_color_stats}
\end{table}

\subsection{Caption Analysis}

Given the intrinsic link between the semantic content of a video and its expressed emotion, we conducted a series of analyses on the captions associated with each video. As a first step, we employed the Natural Language Toolkit (NLTK) to assess the sentiment polarity of each caption. Prior to analysis, all captions were tokenized after removing URLs, numbers, punctuation marks, and standard English stopwords. Figure~\ref{fig:pos/neg} presents the proportion of positive and negative captions for each emotion category. The results reveal a clear pattern: videos labeled with positive emotions tend to have a significantly higher proportion of positively valenced captions compared to those labeled with negative emotions.

Beyond sentiment polarity, we also performed a keyword analysis to uncover representative semantic patterns across different emotional categories. Using the CountVectorizer with n-grams ranging from 2 to 4 words, we extracted the most frequent short phrases for each emotion class. Table~\ref{tab:top_keywords} lists the top five frequent caption-level phrases corresponding to both positive and negative emotions. Interestingly, the extracted keywords reflect the affective semantics of each category. For instance, positive emotions such as contentment and amusement are frequently associated with calm, festive, or joyful expressions (e.g., ``Peace written,'' ``Merry Christmas''), while negative emotions like fear and anger feature more intense or threatening phrases (e.g., ``Man holding gun aiming,'' ``Boxing glove'').

\begin{table*}[t]
\centering
\begin{tabular}{l l l l}
\toprule
\textbf{Contentment} & \textbf{Awe} & \textbf{Amusement} & \textbf{Excitement} \\
\midrule
Brown hugging        & Wonder woman        & Merry Christmas     & Adventure Displayed \\
Peace written        & Addressing group    & Christmas card      & Excited written \\
Expect it            & Enjoy your          & Students seated     & Your enthusiasm \\
Stone lantern        & Pom poms            & Premier league      & Running through lush \\
Blonde boy           & My pleasure         & Hop dress           & Going adventure \\
\midrule
\textbf{Sadness}     & \textbf{Disgust}    & \textbf{Fear}       & \textbf{Anger} \\
\midrule
Blue tears           & Text hate           & Room person standing center & Intense emotion \\
Word pathetic        & Toilet bowl         & Man holding gun aiming     & Shows man flaming \\
Desperate attention  & You disgust me      & Swimming body              & Red angry character \\
Background yellow and black & Shows grinch & Underground tunnel         & Man flaming skill \\
People who desperate & Reads hate          & Bubbles rising             & Boxing glove \\
\bottomrule
\end{tabular}
\caption{\textbf{Top-5 phrases for the caption of each emotion.} The table presents the most frequent short phrases associated with each emotion category, extracted from video captions.}
\label{tab:top_keywords}
\end{table*}

\section{Construction Details}
To ensure high-quality emotion labels across diverse video types, we employed a hybrid labeling strategy that combines manual annotation with classifier-assisted verification.

\subsection{Manual Annotation Protocol}
To retrieve emotion-specific stickers, we used the eight emotion categories along with their synonyms given by EmoSet. Based on this, we further merged words with the same root---for example, amusement/amusing/amused were all mapped to amuse. The full list of keywords is in Table~\ref{tab:emotion_words}.

We recruited a total of 12 annotators to perform fine-grained emotion labeling. Among them, two annotators were tasked with validating the pre-labeled emotion sticker dataset. They reviewed each clip and retained only those whose emotional expressions were consistent with the assigned label. After this filtering process, 6,696 out of the original 9,633 clips were preserved.

The remaining 10 annotators focused on 20\% of the animation and movie datasets—namely, 3,000 clips from movies and 600 from animation. Each clip was independently labeled by three annotators, who were asked to select one category from eight emotions plus a ``no specific emotion'' option. A clip was retained only if at least two annotators reached consensus on the same label. Given the critical role of audio in emotional expression, annotators watched each clip with its audio during the labeling process. 

\subsection{Classifier Comparison}

\begin{figure}[t]
    \centering
    \includegraphics[width=0.9\linewidth]{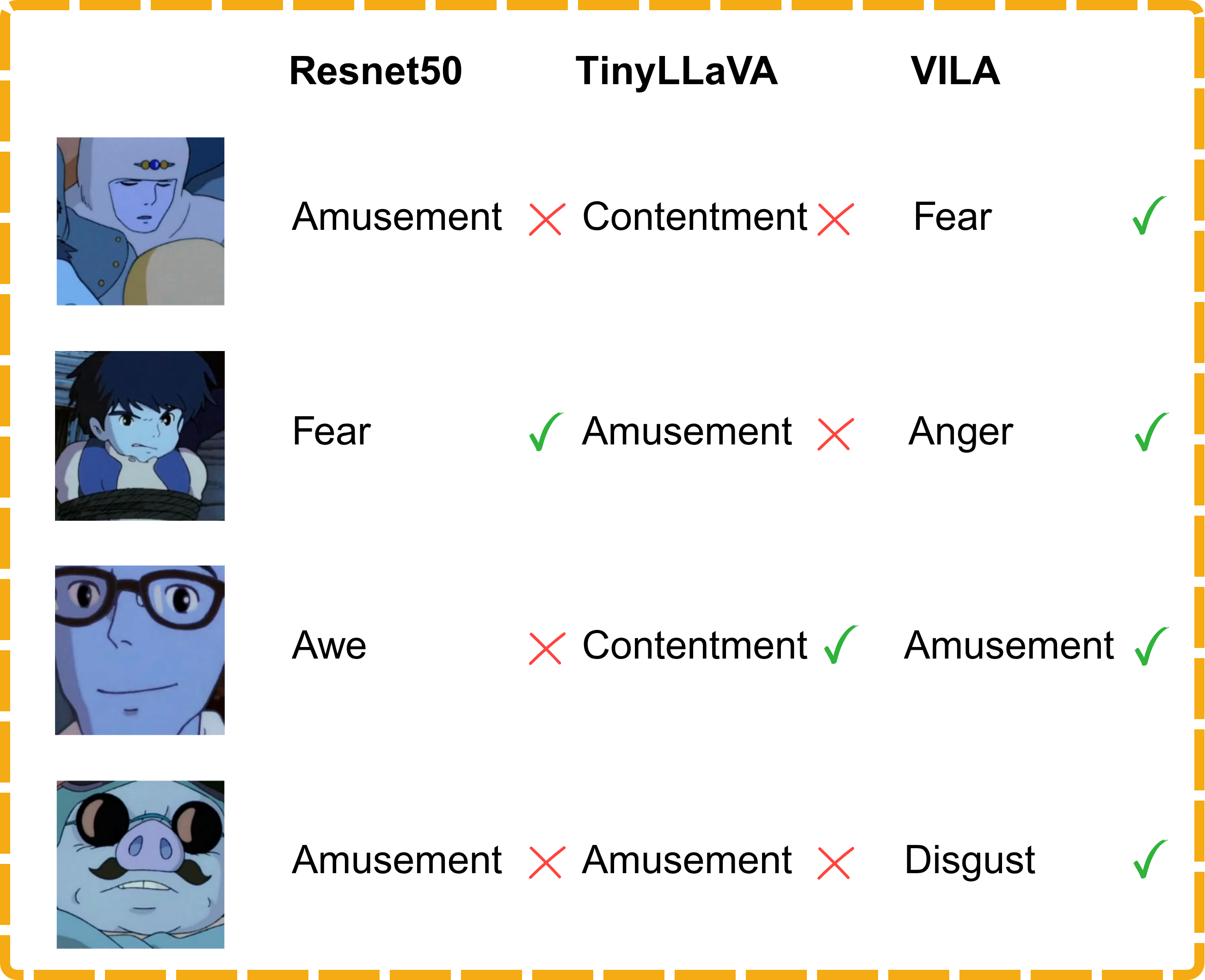}
    \caption{\textbf{Qualitative comparison of automatic labelers on animation samples.} For each clip, we show the predictions from VGG-16, ResNet-50, TinyLLaVA-Phi-2-SigLIP-3.1B, and NVILA-Lite-2B. NVILA-Lite-2B produces the most semantically coherent labels.}
    \label{fig:classifier_comparison}
\end{figure}

To further scale up the labeling process, we evaluate both traditional visual classifiers and recent Vision-Language Models (VLMs) on EmoSet-118k. Specifically, we compared two convolutional baselines—ResNet-50 and VGG-16—with two lightweight VLM pipelines: TinyLLaVA-Phi-2-SigLIP-3.1B and NVILA-Lite-2B. All models were trained on a randomly selected 80\% subset of the EmoSet-118k dataset and evaluated on a held-out set of 160 images. Detailed results are in the Table~\ref{fig:classifier_comparison}.

Based on a small-scale human evaluation, the fine-tuned NVILA-Lite-2B model achieved an accuracy of 87.5\% on EmoSet-118k, closely matching human-level performance. Notably, its misclassifications typically involved semantically adjacent emotions (e.g., \textit{awe} vs. \textit{contentment}), which are inherently difficult to distinguish. In addition, NVILA-Lite-2B exhibited a low computational footprint, making it a practical choice for scalable labeling. Figure~\ref{fig:classifier_comparison} showcases examples from the EmoVid animation subset, highlighting that NVILA-Lite-2B consistently produces semantically coherent emotion labels---often outperforming the other baselines by a significant margin.
\begin{table*}[h]
\centering
\renewcommand{\arraystretch}{1.2} 
\begin{tabular}{lcccccc}
\hline
\textbf{Model} & \textbf{ResNet50} & \textbf{VGG-16} & \textbf{TinyLLaVA} & \textbf{TinyLLaVA (finetuned)} & \textbf{NVILA} & \textbf{NVILA (finetuned)} \\
\hline
\textbf{Accuracy (\%)} & 66.68 & 45.46 & 26.88 & 38.75 & 57.5 & 87.5 \\
\hline
\end{tabular}
\caption{\textbf{Emotion classification accuracy of different models on EmoSet.} TinyLLaVA refers to TinyLLaVA-Phi-2-SigLIP-3.1B, and NVILA refers to NVILA-Lite-2B.}
\label{tab:emotionset}
\end{table*}
To further assess the labeling performance of the VLM on EmoVid, we randomly selected 1\% of the VLM-labeled videos as a validation set. Three human annotators were invited to independently annotate these videos. The overall Fleiss' kappa across all annotations was 0.3704. We also computed pairwise Cohen's kappa scores among the four annotation sources (three humans and the VLM). The average inter-human kappa was 0.311, while the average human-VLM kappa was 0.301, indicating a minimal gap. These results suggest that the VLM achieves human-comparable annotation ability on the EmoVid dataset.

\begin{figure*}
    \centering
    \includegraphics[width=1\linewidth]{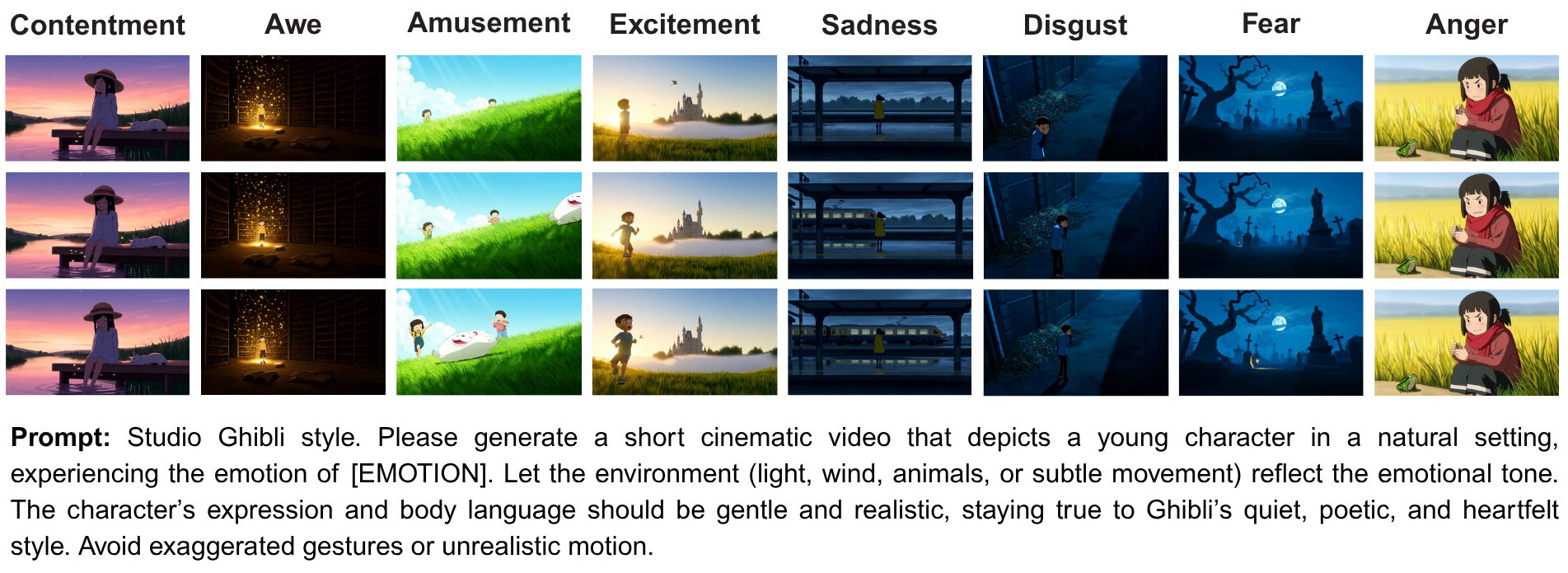}
    \caption{\textbf{Examples of generated videos using our fine-tuned Wan2.1-T2V model.} An extra LoRA module is included to generate Studio Ghibli style videos.}
    \label{fig:t2v}
\end{figure*}

\section{Metrics and Implementation Details}
\label{app:metrics}

\noindent\textbf{Notation.}
Let a generated video be $V=\{x_t\}_{t=1}^{T}$ with $T$ frames, and let $y$ denote the corresponding text prompt. We report \textbf{FVD}$\downarrow$, \textbf{CLIP}$\uparrow$, \textbf{SD}$\uparrow$, \textbf{Flicker}$\downarrow$, and \textbf{EA}$\uparrow$.

\vspace{0.5em}
\noindent\textbf{CLIP score (text--video alignment, $\uparrow$).}
We use OpenCLIP with \emph{ViT-L/14} (pretrained=\texttt{openai}) and its default preprocessing.
For each video, we uniformly sample up to $K=\min(8,T)$ frames and encode them with the image encoder to unit-normalized vectors $\{\mathbf{f}_t\}_{t=1}^{K}$.
The video embedding is the average of per-frame features,
\[
\bar{\mathbf{f}}=\frac{1}{K}\sum_{t=1}^{K}\mathbf{f}_{t}.
\]
The prompt $y$ is tokenized and encoded to a unit vector $\mathbf{g}$.
The CLIP score is the cosine similarity
\[
\mathrm{CLIP}(V,y)=\cos\!\big(\bar{\mathbf{f}},\,\mathbf{g}\big).
\]
We report the mean over the evaluation set.

\vspace{0.5em}
\noindent\textbf{SD score (first-frame consistency in latent space, $\uparrow$).}
This metric measures how consistent the content of all frames is with the first frame in a strong image latent space.
We use the VAE encoder from \texttt{stabilityai/sd-vae-ft-mse} (Diffusers).
Each frame is resized to $512{\times}512$ and normalized to $[-1,1]$.
Let $\mathbf{z}_t$ denote the flattened latent sampled from the encoder's posterior for frame $x_t$; we compute
\[
\mathrm{SD}(V)=\frac{1}{T}\sum_{t=1}^{T}\cos\!\big(\mathbf{z}_1,\,\mathbf{z}_t\big),
\]
and report the mean across videos.

\vspace{0.5em}
\noindent\textbf{Temporal Flicker (perceptual instability, $\downarrow$).}
We quantify frame-to-frame perceptual change using LPIPS with the \texttt{alex} backbone.
All frames are center-cropped to $224{\times}224$ and normalized to $[-1,1]$.
For consecutive frames $(x_t,x_{t+1})$,
\[
\mathrm{Flicker}(V)=\frac{1}{T-1}\sum_{t=1}^{T-1}\mathrm{LPIPS}\!\big(x_t,\,x_{t+1}\big).
\]
Lower is better; we report the average across videos.

\vspace{0.5em}
\noindent\textbf{Content-debiased Fr\'echet Video Distance (FVD, $\downarrow$).}
We follow the content-debiased FVD implementation and extract video features with either VideoMAE-v2 or I3D (as configured in our code).
Given Gaussian fits to the real and generated feature sets, $\mathcal{N}(\mu_r,\Sigma_r)$ and $\mathcal{N}(\mu_g,\Sigma_g)$, 
\[
\mathrm{FVD}=\|\mu_r-\mu_g\|_2^2+\operatorname{Tr}\!\Big(\Sigma_r+\Sigma_g-2\big(\Sigma_r^{1/2}\,\Sigma_g\,\Sigma_r^{1/2}\big)^{1/2}\Big).
\]
Real-set statistics can be cached; generated features are computed on-the-fly.

\vspace{0.75em}
\noindent\textbf{Emotion Accuracy (EA, $\uparrow$).}
We evaluate whether the predicted emotion of a generated video matches the ground-truth (GT) label.

\paragraph{EA\_8cls (8-way accuracy).}
Let $\hat{e}_i\in\mathcal{E}$ and $e_i\in\mathcal{E}$ be the predicted and GT emotion for sample $i$, where $|\mathcal{E}|=8$.
The 8-class top-1 accuracy is
\[
\mathrm{EA\_8cls}=\frac{1}{N}\sum_{i=1}^{N}\mathbb{1}\!\left[\hat{e}_i=e_i\right].
\]

\paragraph{EA\_2cls (valence accuracy).}
We partition the eight emotions into a \emph{positive} set $\mathcal{P}$ and a \emph{negative} set $\mathcal{N}$ with $\mathcal{P}\cup\mathcal{N}=\mathcal{E}$ and $\mathcal{P}\cap\mathcal{N}=\varnothing$ (four classes per side, fixed by our annotation protocol).
Let $\nu:\mathcal{E}\to\{\mathrm{pos},\mathrm{neg}\}$ be the induced valence mapping:
\[
\mathrm{EA\_2cls}=\frac{1}{N}\sum_{i=1}^{N}\mathbb{1}\!\left[\nu(\hat{e}_i)=\nu(e_i)\right].
\]
Intuitively, EA\_2cls counts a prediction as correct whenever the predicted emotion has the same polarity (positive vs.\ negative) as the ground truth.

\vspace{0.5em}
\noindent\textbf{Implementation notes.}
Videos are decoded with \texttt{decord}.
For CLIP, we uniformly sample up to 8 frames; SD and Flicker use all frames.
All cosine similarities and means are computed per video and averaged over the benchmark set.

\section{More Experiment Results}
\begin{figure*}
    \centering
    \includegraphics[width=1\linewidth]{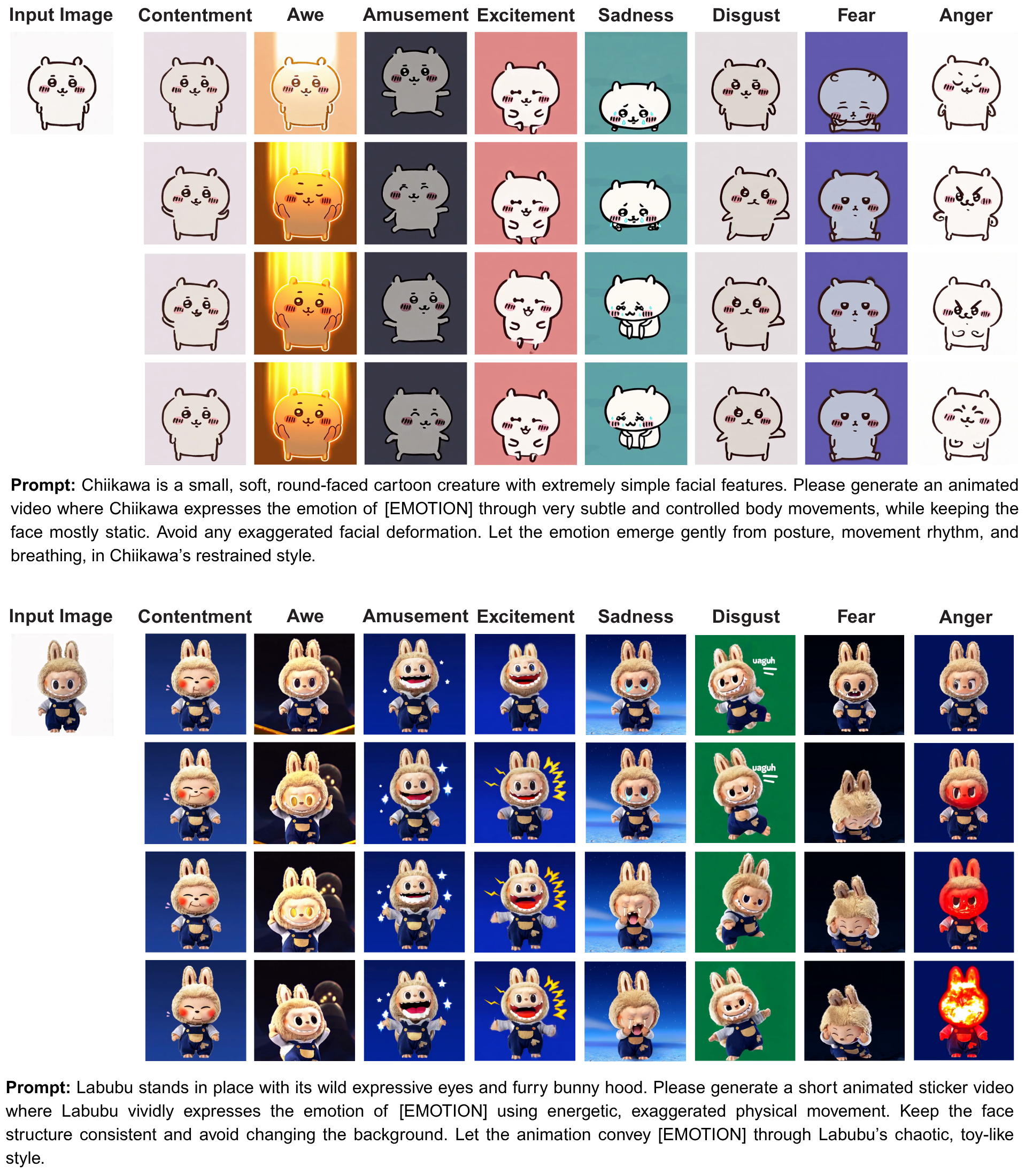}
    \caption{\textbf{Example results of emotion-conditioned animated sticker generation using our fine-tuned Wan2.1-I2V model.} We demonstrate EmoVid's potential in efficiently adapting existing general-purpose video models for emotional content generation tailored to stylized, creative, and social media applications.}
    \label{fig:i2v}
\end{figure*}

We used the fine-tuned Wan2.1-I2V model to generate animated expressions of characters with specific emotions, as shown in Figure~\ref{fig:i2v}. We then applied the Multiple LoRA technique with the Wan2.1-T2V model to generate videos that combine specific visual styles with targeted emotional expressions, as shown in Figure~\ref{fig:t2v}.

\section{User Study}
\begin{figure*}[t]
    \centering
    \includegraphics[width=\linewidth]{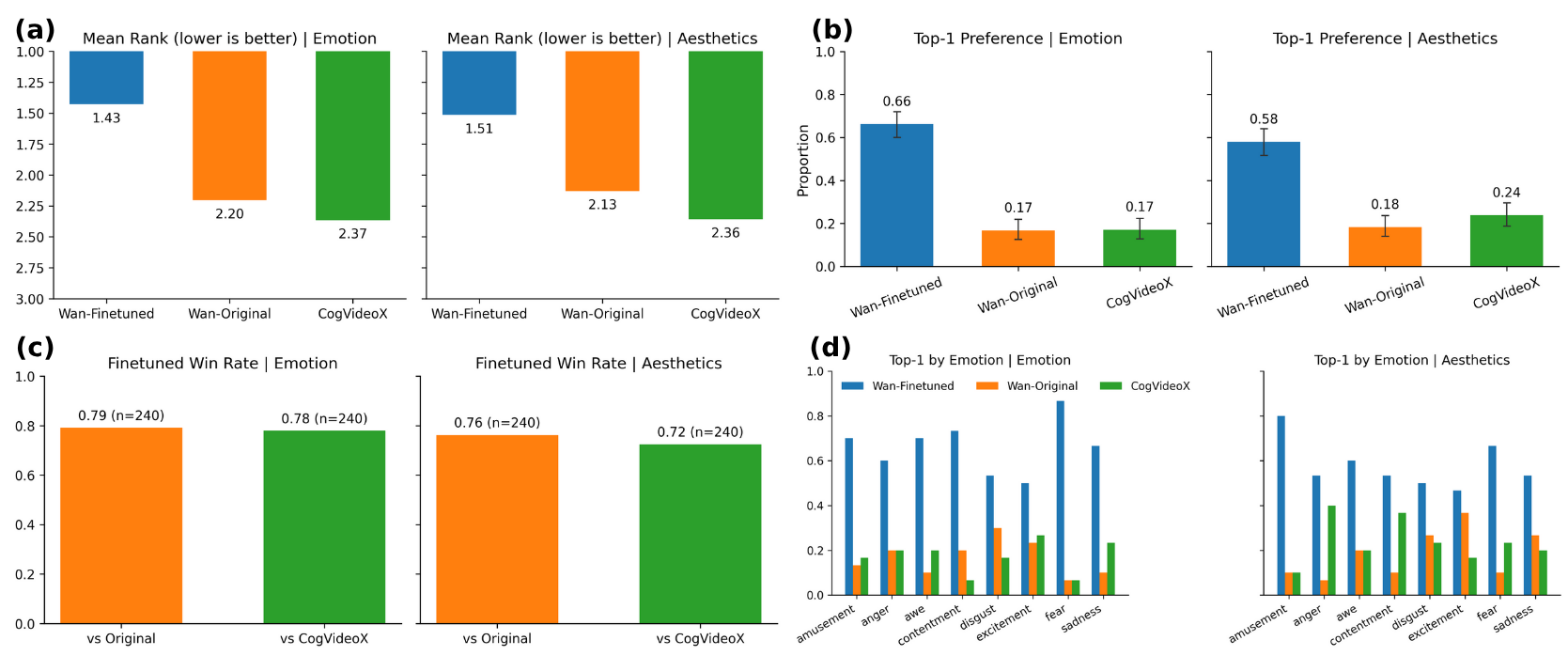}
    \caption{\textbf{Perceptual user study results comparing \texttt{Wan-Finetuned} with \texttt{Wan-Original} and \texttt{CogVideoX}.}
        (\textbf{a}) Mean rank for Emotion Expression and Aesthetic Quality (lower is better).
        (\textbf{b}) Top-1 preference rate (higher is better).
        (\textbf{c}) Pairwise win rate of \texttt{Wan-Finetuned} against the two baselines.
        (\textbf{d}) A per-category breakdown of Top-1 preference rates, demonstrating consistent advantages for \texttt{Wan-Finetuned} across all eight emotions.}
    \label{fig:user_study}
\end{figure*}
We conducted a controlled perceptual study to compare our \texttt{Wan-Finetuned} model against two baselines, \texttt{Wan-Original} and \texttt{CogVideoX}. We generated videos for eight emotion categories, using both text-to-video (T2V) and image-to-video (I2V) prompts for each. Fifteen participants were presented with the outputs from all three models side-by-side and asked to rank them on two criteria: \textbf{Emotion Expression} (the accuracy and salience of the emotion) and \textbf{Aesthetic Quality} (coherence, plausibility, and visual appeal). This process yielded 240 discrete rankings per criterion (15 raters $\times$ 16 prompts).

We analyzed the rankings using mean rank, Top-1 preference rate, and pairwise win rates. We also performed statistical significance tests on the comparisons and computed inter-rater reliability using Kendall's $W$ to validate findings.

The results, summarized in Figure~\ref{fig:user_study}, show a clear and significant preference for \texttt{Wan-Finetuned}. For Emotion Expression, participants ranked \texttt{Wan-Finetuned} first in 66.2\% of comparisons, far exceeding \texttt{Wan-Original} (16.7\%) and \texttt{CogVideoX} (17.1\%) (Fig.~\ref{fig:user_study}b). This strong preference is mirrored in the mean rank (Fig.~\ref{fig:user_study}a) and the decisive pairwise win rates against both baselines (Fig.~\ref{fig:user_study}c). We observed a similar strong preference for \texttt{Wan-Finetuned} on Aesthetic Quality (57.9\% Top-1 rate). As shown in Fig.~\ref{fig:user_study}d, this advantage is consistent across all individual emotion categories. All pairwise preferences for \texttt{Wan-Finetuned} were statistically significant ($p \ll 0.001$), and moderate inter-rater agreement (mean Kendall's $W = 0.371$ for Emotion, $0.333$ for Aesthetics) confirms the reliability of these judgments.

\begin{table*}[t]
\centering
\begin{tabular}{lp{0.8\textwidth}r}
\toprule
\textbf{Emotion} & \textbf{Keyword} \\
\midrule
Amusement & amuse, entertain, delight, enjoy, pleasure, laughter, mirth, hilarity, merry, glad, recreation, extravaganza, cheer, delectation, ravishment  \\
\midrule
Awe & awe, wonder, revere, venerate, inspire, respect, hallowed, exalt, amaze, astonish, impress, marvel, astound, startle, surprise, worship\\
\midrule

Contentment & content, happy, satisfy, fulfill, need, expect, long, complacence, smug, gloat, peace, ease, comfort, gratify, serenity, equanimity, replete, warmth\\
\midrule
Excitement & excite, pride, glee, exhilarate, fervor, lively, joy, rouse, agitate, passion, thrill, adventure, enthusiasm, flurry, furore, commotion, elate, kick, nightlife, show, frisson, hysteria\\
\midrule
Anger & anger, choler, ire, grievance, fury, rage, wrath, infuriate, enrage, umbrage, offend, indignation, outrage, dudgeon, irascible, annoy, chafe, vex, pique, irritate, aggravate, exasperate, harass, torment, displease, resent, antagonize, provoke, hassle, burn, explode, fume, seethe, aggressive\\
\midrule
Disgust & disgust, dislike, distaste, flush, revolt, revulsion, vomit, repel, sicken, abhor, abomination, detest, execration, loathe, odium, repugnance, repulse, nausea, hate, averse, antipathy \\
\midrule
Fear & fear, horror, afraid, scare, frighten, panic, terror, affright, dread, anxious, apprehensive, alarm, dismay, consternation, shiver, chill, quiver, shudder, worry, concern, trouble, uneasy, tremor, qualm, trepidation, timid, craven, funk, creep, attack, intimidate\\
\midrule
Sadness & sad, depress, sorrow, unhappy, bitter, doleful, mourn, melancholy, pensive, wistful, tired, despair, desperate, deplore, distress, lament, pity, sorry, gloom, grieve, dismal, sombre, glum, deject, downcast, tear, lugubrious, disconsolate, cheerless, lachrymose, woebegone, triste, tragic, upset, disastrous, pathetic, poignant, harrow, miserable, heartbreak, wretched, desolate, dispirit\\
\bottomrule
\end{tabular}
\caption{\textbf{Emotion categories and keywords.} Curated keyword lists (lemmas and common synonyms) used for retrieval and normalization during data collection.}
\label{tab:emotion_words}
\end{table*}

\end{document}